\documentclass[twocolumn,a4paper]{IEEEtran} 

\usepackage{amsmath} 
\usepackage{amssymb} 
\usepackage{amsfonts}
\usepackage{color}

\usepackage{cite}
\usepackage{textcomp}

\usepackage{ifpdf}

\ifpdf
   \usepackage[pdftex]{graphicx}
   \graphicspath{{figures/}}
  \DeclareGraphicsExtensions{.pdf,.jpeg,.png,.eps}
\else
\fi
\newcommand{\vc}[1]{{\bf #1}}

\newcommand{\ma}[1]{{\bf #1}}

\newcommand{\rev}[1]{{#1}}

\usepackage{todonotes}

\title{When Slepian Meets Fiedler:\\ Putting a Focus on the Graph Spectrum}

\author{Dimitri Van De Ville, \emph{Senior Member}, Robin Demesmaeker, Maria Giulia Preti, \emph{Member}
\thanks{The authors are with the Institute of Bioengineering/Center for Neuroprosthetics, \'Ecole Polytechnique F\'ed\'erale de Lausanne (EPFL), Lausanne, Switzerland, and the Department of Radiology and Medical Informatics, University of Geneva, Geneva, Switzerland. Corresponding author: D. Van De Ville (e-mail: dimitri.vandeville@epfl.ch).}}

\begin{document}

\maketitle


\begin{abstract}
The study of complex systems benefits from graph models and their analysis. In particular, the eigendecomposition of the graph Laplacian lets emerge properties of global organization from local interactions; e.g., the Fiedler vector has the smallest non-zero eigenvalue and plays a key role for graph clustering. Graph signal processing focusses on the analysis of signals that are attributed to the graph nodes. The eigendecomposition of the graph Laplacian allows to define the graph Fourier transform and extend conventional signal-processing operations to graphs. Here, we introduce the design of Slepian graph signals, by maximizing energy concentration in a predefined subgraph for a graph spectral bandlimit. We establish a novel link with classical Laplacian embedding and graph clustering, which provides a meaning to localized graph frequencies.
\end{abstract}

\section{Introduction}

\IEEEPARstart{G}{raphs} 
are powerful mathematical models to represent complex data structures~\cite{Newman.2010}. Spectral graph theory supports an important class of methods for studying graph topology as well as analyzing graph signals. Graph partitioning problems~\cite{Luxburg.2007} can be solved as the convex relaxation of the graph-cut criterion, which reverts to an eigendecomposition of the graph Laplacian. Thresholding the eigenvector with the smallest non-zero eigenvalue---commonly called the Fiedler vector---leads to a graph bipartition~\cite{Fiedler.1989}. These eigenvectors also solve the embedding problem; i.e., mapping the nodes onto a line (or higher-dimensional space) such that the distance between connected nodes is minimized~\cite{Belkin.2003}. In addition, the same eigenvectors are also used to define the graph Fourier transform (GFT), which has attracted a lot of attention from the signal processing community as many Fourier-domain operations have been generalized to graphs using the GFT~\cite{Shuman.2013}. One notable example is the spectral graph wavelet transform~\cite{Hammond.2011} that defines graph-domain wavelets by using a window function in the spectral domain. 
 
In this Letter, we propose the graph extension of Slepian functions, which were introduced on regular domains by Slepian and colleagues~\cite{Slepian.1961,Slepian.1978}, to find a trade-off between temporal and spectral energy concentration. 
These functions have been extended to several other domains, including spherical ones in the context of geophysics~\cite{Simons.2006}. Balancing the spread of graph signals in the original and the spectral domain relates to extensions of uncertainty principles~\cite{Agaskar.2013}, which have been explored by Tsitsvero \emph{et al.}~\cite{Tsitsvero.2016} based on a Slepian-type of constraints. \rev{Here, we take the design of graph Slepians one step further by establishing a link between the criterion of energy concentration and the one of Laplacian embedding and graph clustering. This leads to Slepians that are optimized w.r.t.~a modified embedded distance, and that are particularly interesting to generalize graph signal processing operations.} 

\section{Defining Slepian Functions on Graphs}

\subsection{Slepian's Time-Frequency Concentration Problem}
\label{subsec:1d}
In seminal work, Slepian, Landau, and Pollak~\cite{Slepian.1961,Slepian.1978} solved the fundamental problem of optimally concentrating a signal jointly in temporal and spectral domains. Here we briefly review the 1-D discrete-discrete case, which is directly relevant for the generalization that we will introduce in this work. We define the unitary discrete Fourier transform (DFT) for a length-$N$ signal by the $N\times N$ matrix 
$$
   \ma{F}= \frac{1}{\sqrt{N}} [ e^{j\omega_l (k-1)}]_{k,l},\quad  k,l=1,\ldots,N,
$$ 
where $\omega_l=2\pi (l-\lceil (N-1)/2 \rceil)/N$. Then, the signal represented as a vector $\vc{f}$ can be Fourier transformed as $\hat{\vc{f}}=\ma{F}^H \vc{f}$, where $\cdot^H$ indicates the Hermitian transpose. Due to Parseval, we have that $\| \vc{f} \|^2 = \| \hat{\vc{f}} \|^2$, where the $\ell_2$-norm is defined as $\| \vc{f} \|^2=\left< \vc{f}, \vc{f} \right>=\vc{f}^H \vc{f}$.

The problem at hand is now to optimally concentrate a strictly band-limited signal $\vc{g}$ in a predefined interval defined by the set of indices $\mathcal{S}$ that does not necessarily need to be contiguous. Finding such an optimal signal in terms of energy concentration in $\mathcal{S}$ can be formulated as maximizing 
\begin{eqnarray}
  \label{eq:slepian-1d-1}
  \mu = \frac{ \sum_{k\in \mathcal{S}} |g[k]|^2 }{ \sum_{k=1}^{N} |g[k]|^2}.
\end{eqnarray}
By definition, the band-limited signal can be represented by a linear combination of a truncated Fourier basis; i.e., we write $\vc{g}=\ma{F} \ma{W} \hat{\vc{g}}$, where $\ma{W}$ of size $N\times N_W$ selects the first $N_W$ Fourier vectors with smallest $|\omega_l|$. In addition, we introduce the $N\times N$ diagonal selection matrix $\ma{S}$ with elements $S_{k,k}=0/1$ that indicate the presence of index $k$ in $\mathcal{S}$; so the number of selected indexes is $\text{trace}(\ma{S})=N_S$. This allows us now to rewrite Eq.~(\ref{eq:slepian-1d-1}) as the Rayleigh quotient
\begin{eqnarray}
  \label{eq:slepian-1d-Rayleigh}
   \mu = \frac{ \hat{\vc g}^H \ma C \hat{\vc g}}{\hat{\vc g}^H \hat{\vc g}},
\end{eqnarray}
where $\ma{C}=\ma{W}^T \ma{F}^H  \ma{S} \ma{F} \ma{W}$ is the concentration matrix of size $N_W\times N_W$. Finding the optimal solution translates into an eigendecomposition problem, $\ma C\hat{\vc s}_k=\mu_k \hat{\vc s}_k$, where $\mu_k$ represents the energy concentration of the Slepian vector $\vc s_k=\ma F \ma{W} \hat{\vc s}_k$ in $\mathcal{S}$. 

\begin{figure}[t]
  \centering
  \begin{tabular}{c}
    \includegraphics[width=8.4cm]{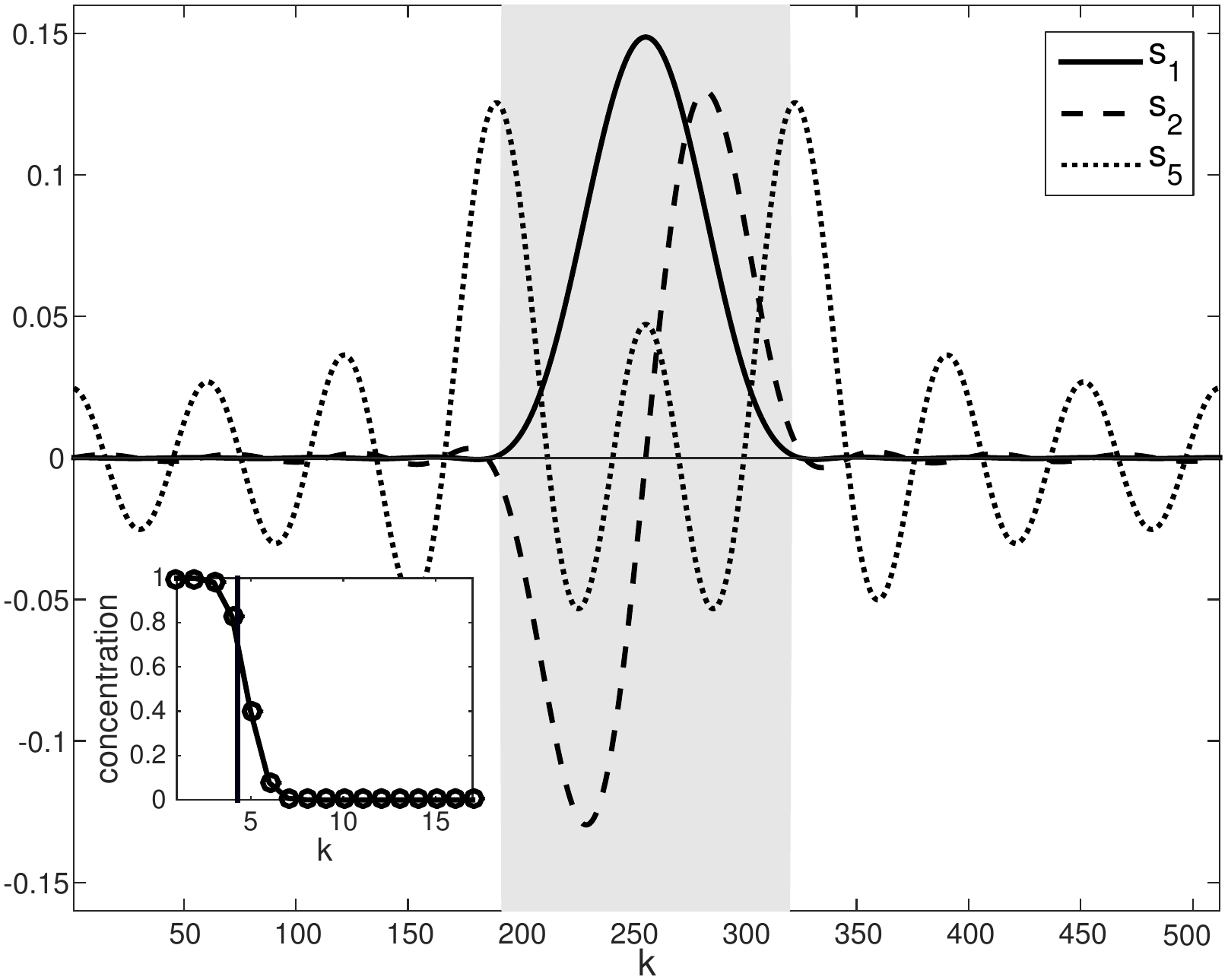}
  \end{tabular}
  \caption{\label{fig:1d} Example of 1-D Slepian vectors for a signal of length $N=512$. Three example vectors are shown for the selected interval indicated with the gray background. The energy concentration is plotted in the inset and the Shannon number is marked by the vertical bar.}
\end{figure}

The Slepian vectors are illustrated for a signal of length $N=512$. The selected interval is chosen centered at $k=256$ with length $N_S=129$. The bandlimited space contains $N_W=17$ Fourier basis vectors ($\left|\omega_l\right|\le \pi/32$). In Fig.~\ref{fig:1d}, the first Slepian vector $\vc{s}_1$ has a Gaussian shape and is well localized in the interval. The second Slepian vector $\vc{s}_2$ is orthogonal to the first one (both over the original domain and the selected interval as shown above) and still well localized. We then show $\vc{s}_5$ which has a weaker energy concentration in the interval. 
By convention, the eigenvalues of the Slepian eigendecomposition are sorted according to decreasing energy concentration ${1>\mu_1\ge \mu_2 \ge \ldots > 0}$ in $\mathcal{S}$. The eigenvalue spectrum presents a sudden transition between well-localized and poorly-localized eigenvectors; this phase transition has been well studied and occurs at the time-bandwidth product or Shannon number $K=N_W N_S/N$. The inset of Fig.~\ref{fig:1d} illustrates this phenomenon which happens around $K=4.3$ for this example. 

\subsection{Graph Fourier Spectrum}
For an undirected weighted graph with $N$ nodes, the connectivity is characterized by an $N\times N$ symmetric adjacency matrix $\ma A$, where the elements $A_{i,j}$ indicate the edge weights between nodes $i,j=1,\ldots,N$. Here we assume the graph is connected and the edge weights are positive. Next to the graph and its topology, we can also consider graph signals that define a mapping from the nodes to a length-$N$ vector associating a value with every node; e.g., the graph topology models the presence of communication links between nodes, while the graph signal represents measures taken at the nodes. 

The graph Laplacian is defined as $\ma L'=\ma D-\ma A$, where $\ma D$ is the diagonal degree matrix with elements $D_{i,i}=\sum_{j=1}^N A_{i,j}$. We further consider the normalized graph Laplacian $\ma{L}=\ma{D}^{-1/2}\ma{L}'\ma{D}^{-1/2}$ that factors out differences in degree and thus is only reflecting relative connectivity.

Spectral methods for graphs are based on the insight that the eigendecomposition of the graph Laplacian gives access to a GFT. The eigenvectors $\vc u_k$ of $\ma L$ minimize 
\begin{equation}
  \label{eq:lapl}
  \lambda = \frac{\vc u^T \ma L\vc u}{\vc u^T \vc u}
\end{equation}
and can be ordered according to increasing eigenvalues ${\lambda_1=0\le \lambda_2 \le \ldots \le \lambda_N}$. The eigenvectors play the role of basis vectors of the graph spectrum, and the associated eigenvalues of frequencies~\cite{Chung.1997}.

\subsection{Slepians on Graphs}
To generalize Slepians to graphs, we follow the approach by~\cite{Tsitsvero.2016}, which is based on introducing selectivity and bandwidth. First, selectivity can be specified by a subset $\mathcal{S}$ that contains the $N_S$ nodes in which we want the energy concentration to be optimal. Similar to the 1-D case, we represent the subgraph by the selection matrix $\ma S$. Second, for the notion of ``bandwidth'', we propose to restrict the spectrum to the eigenvectors with the $N_W<N$ smallest eigenvalues $\lambda$. We then consider the truncated graph spectrum matrix $\ma U\ma{W}$ of size $N\times N_W$; i.e., any band-limited graph signal can be represented by $\vc{g}=\ma{U} \ma{W} \hat{\vc{g}}$. 

Finding the linear combination of eigenvectors within the bandlimit $N_W$ and with maximal energy in $\mathcal{S}$ reverts to optimizing the Rayleigh quotient 
\begin{equation}
\label{eq:slepian}
  \mu = \frac{\hat{\vc g}^T \ma C \hat{\vc g}}{\hat{\vc g}^T \hat{\vc g}},
\end{equation}
where $\ma{C}=\ma{W}^T \ma{U}^T \ma{S} \ma{U} \ma{W}$ is the concentration matrix. Since $\ma{L}$ is real and symmetric, $\ma{U}$ is real as well and we revert to the regular transpose $\cdot^T$.
\rev{The Slepian vectors are orthonormal over the entire graph as well as orthogonal over the interval $\mathcal{S}$; i.e., we have $\vc{g}_k^T \vc{g}_l=\delta_{k-l}$ as well as $\vc{g}_k^T \ma{S} \vc{g}_l  = \mu_k \delta_{k-l}$, where $\delta$ is the Kronecker delta.}

\section{Link with Laplacian Embedding and\\ Graph Clustering}
The GFT plays a central role in Laplacian embedding and graph clustering. Specifically, in Laplacian embedding, the aim is to find a mapping from the nodes onto a line such that strongly connected nodes stay as close as possible, which can be expressed as~\cite{Belkin.2003}:
\begin{equation}
  \label{eq:lap_embedding}
  \arg \min_{\vc g} \sum_{i,j=1}^{N} A_{i,j} (g_i - g_j)^2 = \arg \min_{\vc g} \vc{g}^T \ma{L} \vc{g},
\end{equation}
where $\vc{g}^T\vc{g}=1$ and $\vc{g}^T \vc {1}=0$. The solution is the eigenvector of the Laplacian with the smallest non-zero eigenvalue, the Fiedler vector~\cite{Fiedler.1989}. The optimization (\ref{eq:lap_embedding}) is also related to the classical graph cut  problem that consists of partitioning the graph into clusters of nodes such that the cut size is minimal. In this setting, $g_i=\pm 1$ indicate the labels of the nodes, which can be relaxed to take any real values. 

We now use the Slepian design to generalize the criterion~(\ref{eq:lap_embedding}) as to find a bandlimited solution that is optimized for a selected set of nodes. To reveal this link, we first rewrite the quadratic form (\ref{eq:lap_embedding}) as
$
  \vc{g}^T \ma{L} \vc{g} = \vc{g}^T \ma{U} \ma{\Lambda} \ma{U}^T \vc{g} = \hat{\vc g}^T \ma{\Lambda} \hat{\vc g}.
$
By observing the identity $\ma{\Lambda}=\ma{\Lambda}^{1/2} \ma{U}^T \ma{U} \ma{\Lambda}^{1/2}$, we propose  to generalize the criterion by introducing the bandwidth selection in the spectral, and node selection in the graph domain, respectively: 
\begin{equation}
  \label{eq:slepian-emb}
  \xi=\hat{\vc g}^T \ma{\Lambda}^{1/2}_W \underbrace{\ma{W}^T \ma{U}^T \ma{S} \ma{U} \ma{W}}_{\ma{C}} \ma{\Lambda}^{1/2}_W \hat{\vc g},
\end{equation}
\rev{where $\ma{\Lambda}_W=\ma W^T \ma{\Lambda}\ma W$ is the band-limited $N_W\times N_W$ diagonal matrix.} 
\rev{Therefore, we propose to solve the eigendecomposition of the modified concentration matrix $\ma{C}_\text{emb} = \ma{\Lambda}^{1/2}_W \ma{C} \ma{\Lambda}^{1/2}_W$.}
\rev{As before, these Slepian vectors are double orthogonal as we have $\vc{g}_k^T \vc{g}_l=\delta_{k-l}$ and $\vc{g}_k^T \ma{S} \vc{g}_l=\xi_k \delta_{k-l}$.}

This demonstrates that Laplacian embedding can be generalized as a Slepian problem with the additional weighting of the Laplacian eigenvalues. It is interesting to note that the eigenvalues $\xi$ of $\ma{C}_\text{emb}$ represent the modified embedded distance, or, equivalently, a ``frequency'' that is localized in the subgraph $\mathcal{S}$, whereas the eigenvalues $\mu$ of $\ma{C}$ represent the energy concentration in the subgraph.

\begin{figure}[t]
 \centering
 \begin{tabular}{cc}
 \includegraphics[width=3.7cm]{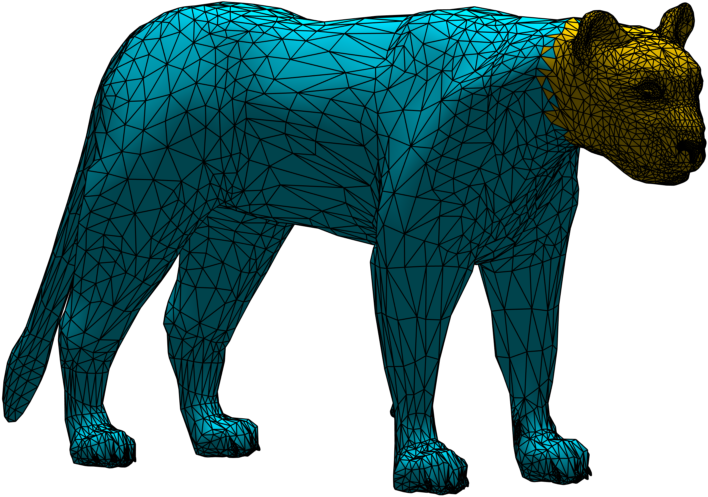} & 
 \includegraphics[width=3.45cm]{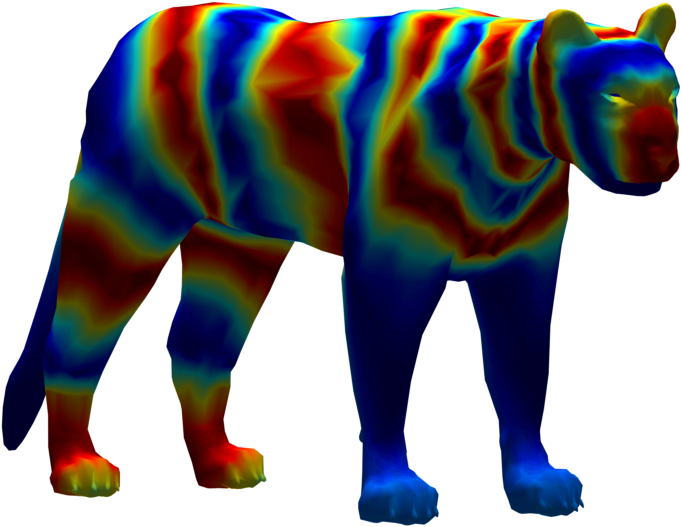} \\
 (a) & (b) \\
\includegraphics[width=3.45cm]{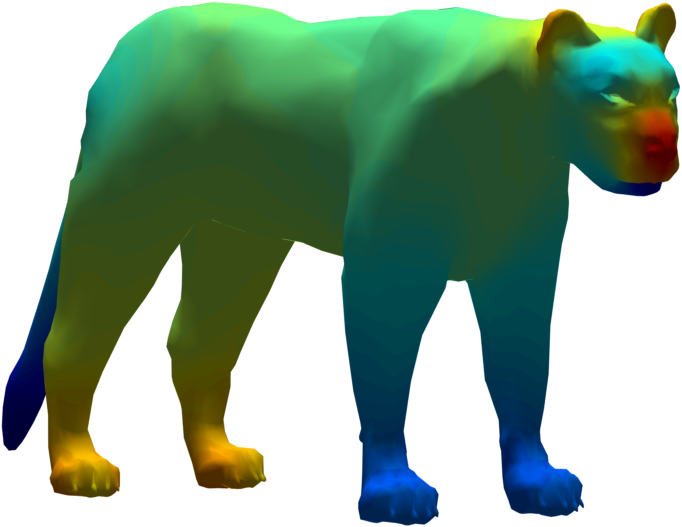} &
\includegraphics[width=3.45cm]{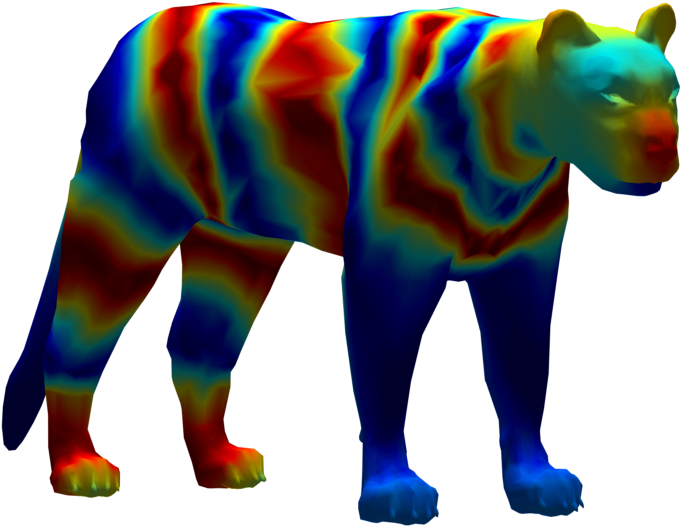} \\
 (c) & (d)
 \end{tabular} \includegraphics[width=0.7cm]{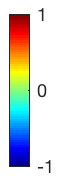}
 \caption{\label{fig:mesh} (a)~Visualization of the mesh that is used to build the graph for the example. The head of the animal is part of the subgraph that is used for the Slepian design. \rev{(b)~Graph signal by taking the sinus of the coordinate of the graph-Laplacian eigenvector with smallest non-zero eigenvalue (i.e., \#2 in Fig.~\ref{fig:mesh-lapl}) such that 8 cycles are obtained. (c)~Filtered version of (b) by applying $\exp(-40\lambda)$ to the graph-Laplacian spectrum. (d)~Filtered version of (b) by applying $\exp(-40\xi)$ to the Slepian spectrum.}}
 \end{figure}

\begin{figure}[t]
\centering
\begin{tabular}{c}
\includegraphics[width=8.7cm]{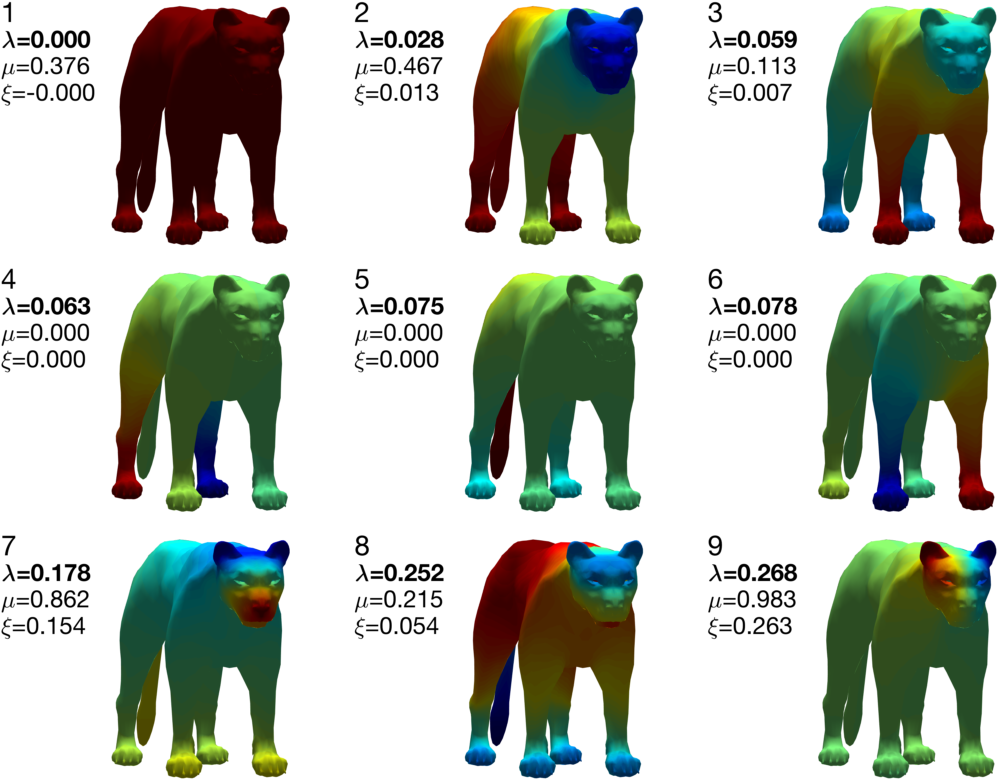}
\end{tabular}
\caption{\label{fig:mesh-lapl} Visualization of first $9$ Laplacian eigenvectors with increasing eigenvalues $\lambda$. The second value ($\mu$) indicates the energy concentration in the subgraph (head) according to Eq.~(\ref{eq:slepian}); the third one ($\xi$) the embedded distance in the subgraph according to Eq.~(\ref{eq:slepian-emb}). \#2 is the Fiedler vector.}
\end{figure}

\begin{figure*}[t]
\centering
\begin{tabular}{ccc}
\includegraphics[width=8.7cm]{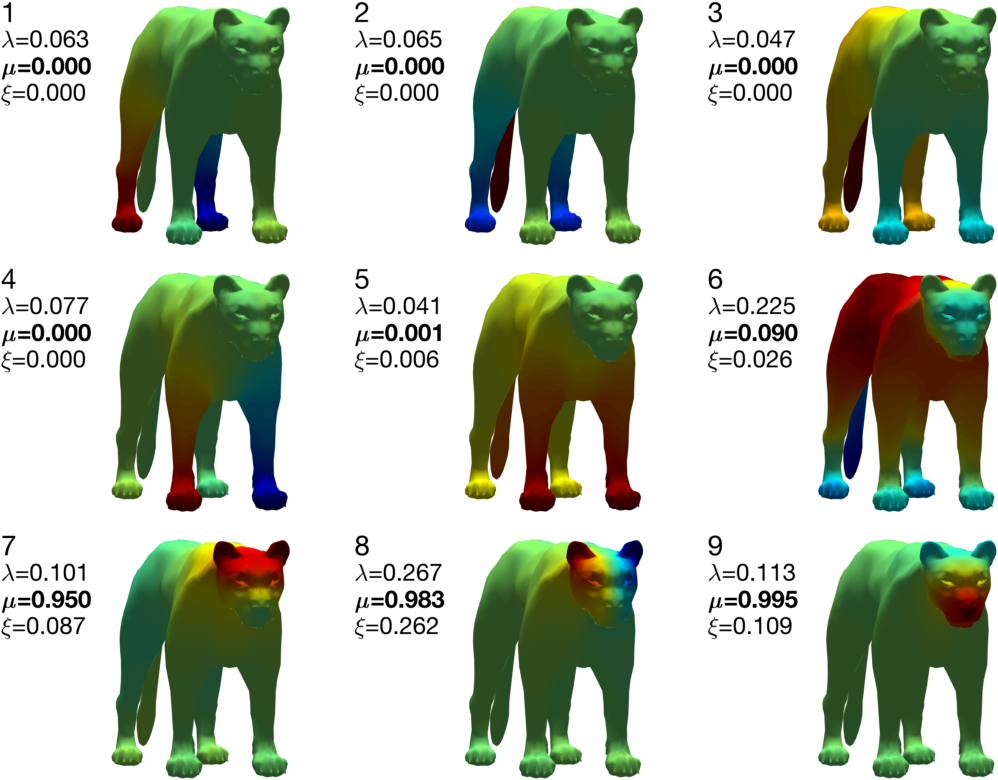} & 
\includegraphics[width=8.7cm]{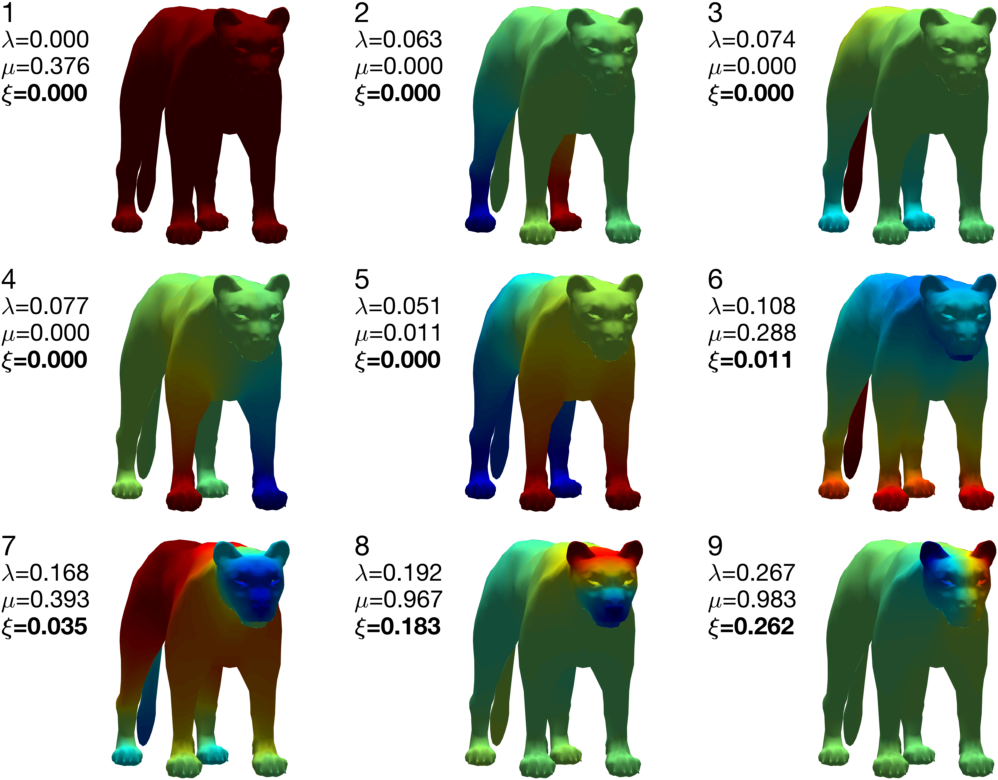} \\
(a) & (b)
\end{tabular}
\caption{\label{fig:mesh-slepian} (a)~Slepian design according to energy concentration in the subgraph with increasing eigenvalues $\mu$. (b)~Slepian design according to embedded distance in the subgraph with increasing eigenvalues $\xi$. The first value ($\lambda$) indicates the conventional Laplacian embedded distance.}
\end{figure*}

\begin{figure*}[t]
\centering
\hspace*{-2ex}
\begin{tabular}{ccc}
\includegraphics[width=5.75cm]{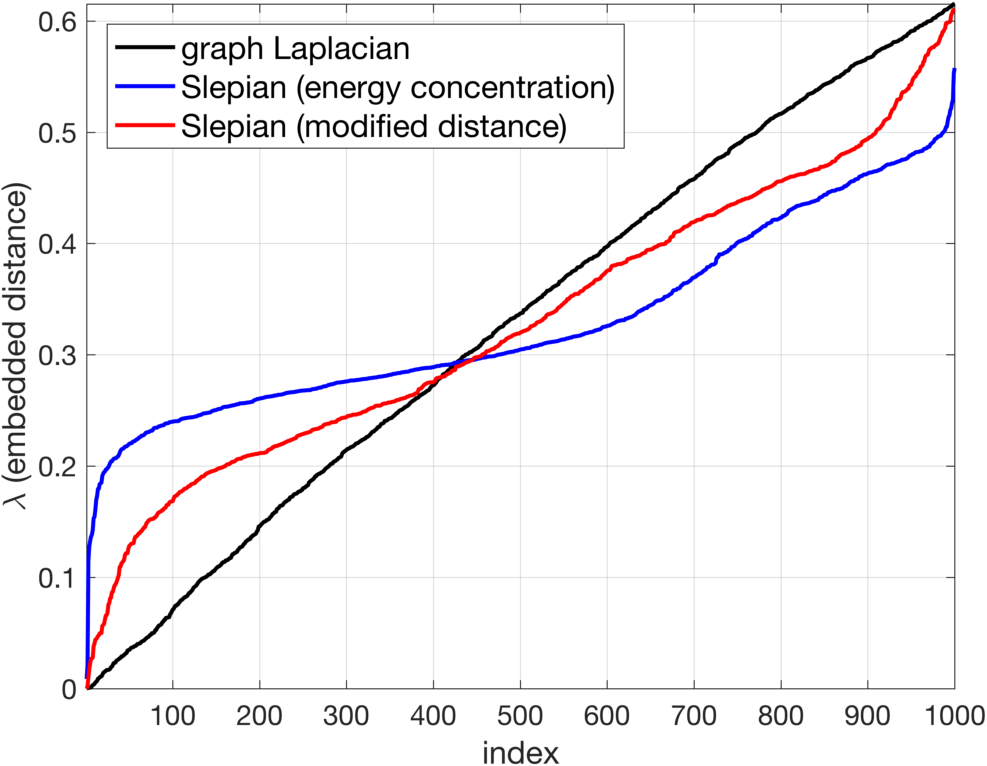} & 
\includegraphics[width=5.75cm]{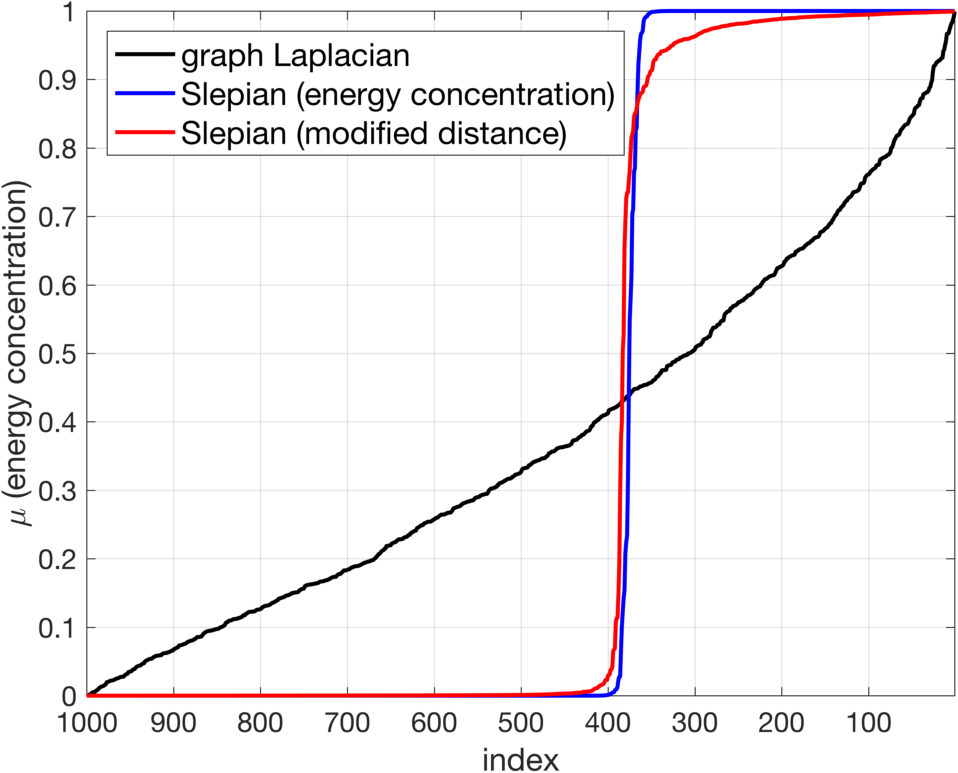} & 
\includegraphics[width=5.75cm]{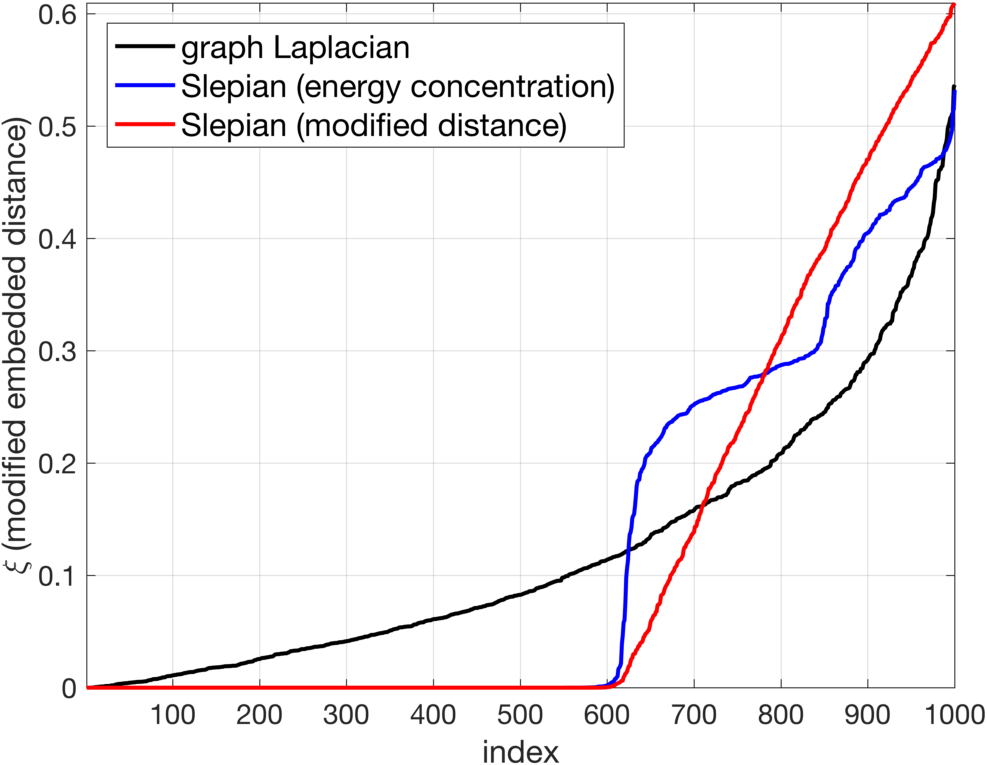} 
\end{tabular}
\caption{\label{fig:plot} \rev{Plots of embedded distance ($\lambda$), energy concentration ($\mu$), and modified embedded distance ($\xi$), for the graph Laplacian and both Slepian designs.}}
\end{figure*}

\section{Results \& Discussion}
\rev{Practically, the design only requires a limited number of eigenvectors $N_W$ that can be obtained using efficient large-scale solvers~\cite{Lehoucq.1996} 
such as ARPACK. The Slepian optimization itself is an eigendecomposition of a matrix of size $N_W\times N_W$.}

We build the graph of a mesh structure with 4'567 nodes and 13'650 edges (9'078 faces), see Fig.~\ref{fig:mesh}a. In Fig.~\ref{fig:mesh-lapl}, the first $9$ Laplacian eigenvectors with smallest eigenvalues are represented. These graph signals correspond to our intuition about low-frequency Fourier basis vectors that represent slow variations along the principle ``axes of variation''.  The Fiedler vector is \#2, which means that 1-D Laplacian embedding projects nodes according to the posterior-anterior axis; i.e., an optimal graph cut separates the front from the back.

We selected a subgraph that corresponds to the head of the animal (i.e., $N_S=1'534$ nodes), see Fig.~\ref{fig:mesh}a. The bandwidth ratio $N_W/N$ is a design choice, which we choose here as $9/4'567=0.2\%$. 
\rev{The Slepian vectors based on the concentration criterion of Eq.~(\ref{eq:slepian}) are illustrated in Fig.~\ref{fig:mesh-slepian}a ordered with increasing eigenvalues; e.g., vector \#9 has highest eigenvalue (0.995) and a strong positive signal in the lower part of the head, in particular, the snout region. 
The vectors based on the modified embedded distance of Eq.~(\ref{eq:slepian-emb}) are shown in Fig.~\ref{fig:mesh-slepian}b. 
For all solutions, we computed the Laplacian embedded distance ($\lambda$), the energy concentration ($\mu$), and the modified embedded distance ($\xi$). The values typeset in bold are true eigenvalues.} 

\rev{We then increased the bandwidth to $N_W=1'000$ and observed how the eigenvalues distribute for the graph Laplacian and both Slepian designs. In Fig.~\ref{fig:plot}, we plotted the corresponding values of $\lambda$, $\mu$, and $\xi$. 
We observe that the eigenvalues of the graph Laplacian are nearly linear in this part of the spectrum. In terms of energy concentration, both Slepian designs exhibit a phase transition known from the 1-D case, approximately at the classical Shannon number $K=N_W N_S/N=336$. The Slepian design based on energy concentration is showing a sharper transition band. Finally, the Slepian design based on the modified embedded distance leads to an almost linear behavior of the eigenvalues $\xi$ for well-concentrated Slepian vectors, which shows a similar behavior of the localized spectrum as for the graph Laplacian.}

\rev{While the Slepian design based on energy concentration is directly relevant for extending uncertainty principles for graphs~\cite{Agaskar.2013,Tsitsvero.2016}, the proposed design based on the modified embedded distance is particularly interesting for graph signal processing since the eigenvalues reflect a localized frequency~\cite{Shuman.2013}. For instance, using again $N_W=1'000$, the graph signal shown in Fig.~\ref{fig:mesh}b is filtered by applying the same window function to the graph spectrum (Fig.~\ref{fig:mesh}c) and to the Slepian spectrum (Fig.~\ref{fig:mesh}d). This demonstrates how the Slepian design can be used to combine spectral operations with controlled localization, and thus are an alternative to graph wavelet approaches~\cite{Hammond.2011,leonardi1302,Tremblay.2014}. Future work can focus on applications including signal recovery and interpolation~\cite{Chen.2015}, adapting the design for directed graphs~\cite{Sandryhaila.2013}, or investigating how the Slepian designs depend on graph regularity.}


\begin{thebibliography}{10}

\bibitem{Newman.2010}
M.~Newman,
\newblock {\em Networks: An Introduction},
\newblock Oxford Univ. Press, New York, NY, USA, 2010.

\bibitem{Luxburg.2007}
U.~Luxburg,
\newblock ``A tutorial on spectral clustering,''
\newblock {\em Statist. Comput.}, vol. 17, no. 4, pp. 395--416, 2007.

\bibitem{Fiedler.1989}
M.~Fiedler,
\newblock ``Laplacian of graphs and algebraic connectivity,''
\newblock {\em Combinatorics and Graph Theory}, vol. 25, pp. 57--70, 1989.

\bibitem{Belkin.2003}
M.~Belkin and P.~Niyogi,
\newblock ``Laplacian eigenmaps for dimensionality reduction and data
  representation,''
\newblock {\em Neural Computation}, vol. 15, no. 6, pp. 1373--1396, 2003.

\bibitem{Shuman.2013}
D.~I. Shuman, S.~K. Narang, P.~Frossard, A.~Ortega, and P.~Vandergheynst,
\newblock ``The emerging field of signal processing on graphs: Extending
  high-dimensional data analysis to networks and other irregular domains,''
\newblock {\em IEEE Signal Processing Magazine}, vol. 30, no. 3, pp. 83--98,
  2013.

\bibitem{Hammond.2011}
D.~K. Hammond, P.~Vandergheynst, and R.~Gribonval,
\newblock ``Wavelets on graphs via spectral graph theory,''
\newblock {\em Appl. and Comp. Harm. Anal.}, vol. 30, no. 2, pp. 129--150,
  2011.

\bibitem{Slepian.1961}
D.~Slepian and H.~O. Pollak,
\newblock ``Prolate spheroidal wave functions, fourier analysis and
  uncertainty,''
\newblock {\em Bell System Technical Journal}, vol. 40, no. 1, pp. 43--63,
  1961.

\bibitem{Slepian.1978}
D.~Slepian,
\newblock ``Prolate spheroidal wave-functions, fourier-analysis, and
  uncertainty. {D}iscrete case,''
\newblock {\em Bell System Technical Journal}, vol. 57, no. 5, pp. 1371--1430,
  1978.

\bibitem{Simons.2006}
F.~J. Simons, F.~A. Dahlen, and M.~A. Wieczorek,
\newblock ``Spatiospectral concentration on a sphere,''
\newblock {\em SIAM Review}, vol. 48, no. 3, pp. 504--536, Jan. 2006.

\bibitem{Agaskar.2013}
A.~Agaskar and Y.~M. Lu,
\newblock ``A spectral graph uncertainty principle,''
\newblock {\em IEEE Transactions on Information Theory}, vol. 59, no. 7, pp.
  4338--4356, 2013.

\bibitem{Tsitsvero.2016}
M.~Tsitsvero, S.~Barbarossa, and P.~Di~Lorenzo,
\newblock ``Signals on graphs: Uncertainty principle and sampling,''
\newblock {\em IEEE Transactions on Signal Processing}, vol. 64, no. 18, pp.
  4845--4860, 2016.

\bibitem{Chung.1997}
F.~R.~K. Chung,
\newblock {\em Spectral Graph Theory},
\newblock Amer. Math. Soc., Providence, 1997.

\bibitem{Lehoucq.1996}
R.~B. Lehoucq and D.~C. Sorensen,
\newblock ``Deflation techniques for an implicitly restarted {A}rnoldi
  iteration,''
\newblock {\em SIAM Journal on Matrix Analysis and Applications}, vol. 17, no.
  4, pp. 789--821, 1996.

\bibitem{leonardi1302}
N.~Leonardi and D.~Van De~Ville,
\newblock ``Tight wavelet frames on multislice graphs,''
\newblock {\em IEEE Transactions on Signal Processing}, vol. 61, no. 13, pp.
  3357--3367, 2013.

\bibitem{Tremblay.2014}
N.~Tremblay and P.~Borgnat,
\newblock ``Graph wavelets for multiscale community mining,''
\newblock {\em IEEE Transactions on Signal Processing}, vol. 62, no. 20, pp.
  5227--5239, 2014.

\bibitem{Chen.2015}
S.~Chen, R.~Varma, A.~Sandryhaila, and J.~Kovacevic,
\newblock ``Discrete signal processing on graphs: Sampling theory,''
\newblock {\em IEEE Transactions on Signal Processing}, vol. 63, no. 24, pp.
  6510--6523, Dec 2015.

\bibitem{Sandryhaila.2013}
A.~Sandryhaila and J.~M.~F. Moura,
\newblock ``Discrete signal processing on graphs,''
\newblock {\em IEEE Transactions on Signal Processing}, vol. 61, no. 7, pp.
  1644--1656, Mar. 2013.

\end{thebibliography}

\end{document}